\documentclass{article}

\usepackage{microtype}
\usepackage{graphicx}
\usepackage{booktabs} 

\usepackage{hyperref}

\usepackage[accepted]{icml2019}

\usepackage[utf8]{inputenc} 
\usepackage[T1]{fontenc}    
\usepackage{url}            
\usepackage{amsfonts}       
\usepackage{nicefrac}       
\usepackage{microtype}      
\usepackage{lipsum}
\usepackage{balance}
\usepackage{makecell}
\usepackage{multirow}
\newtheorem{definition}{Definition}

\begin{document}

\twocolumn[
\icmltitle{Recognizing License Plates in Real-Time}





\begin{icmlauthorlist}
\icmlauthor{Xuewen Yang}{sbu}
\icmlauthor{Xin Wang}{sbu}
\end{icmlauthorlist}

\icmlaffiliation{sbu}{Stony Brook University, Stony Brook, US}
\icmlcorrespondingauthor{Xuewen Yang}{xuewen.yang@stonybrook.edu}

\vskip 0.3in
]




\printAffiliationsAndNotice{}  

\begin{abstract}
License plate detection and recognition (LPDR) is of growing importance for enabling intelligent transportation and ensuring the security and safety of the cities. However, LPDR faces a big challenge in a practical environment.  The license plates can have extremely diverse sizes, fonts and colors, and the plate images are usually of poor quality caused by skewed capturing angles, uneven lighting, occlusion, and blurring.  In applications such as surveillance, it often requires fast processing. To enable real-time and accurate license plate recognition, in this work, we propose a set of techniques: 1) We introduce a {\em contour reconstruction} method along with edge-detection to quickly detect the candidate plates;   2) We design a simple {\em zero-one-alternation} scheme to effectively remove the fake top and bottom borders around plates to facilitate more accurate segmentation of characters on plates; 3) To address the overfitting problem that prevents the use of convolutional neural networks (CNN) for character recognition, we introduce a set of techniques to augment the training data, incorporate SIFT features into the CNN network, and exploit transfer learning to obtain the initial parameters for more effective training; and 4) We take a two-phase verification procedure to determine the correct plate at low cost, a {\em statistical filtering} in the plate detection stage to quickly remove unwanted candidates, and the accurate CR results after the CR process to perform further plate verification without additional processing. We implement a complete LPDR system based on our algorithms. The experimental results demonstrate that our system can accurately recognize license plate in real-time. Additionally, it works robustly under various levels of illumination and noise, and in the presence of car movement. Compared to peer schemes, our system is not only among the most accurate ones but is also the fastest, and can be easily applied to other scenarios.
\end{abstract}

\section{Introduction}

License plate detection and recognition (LPDR) of vehicles has been playing an increasingly important role in building intelligent societies and cities. LPDR can be exploited to enforce the security of communities, enable road safety, and facilitate the collection of payment for parking or road toll \cite{hsu2013application}. Although there are many recent efforts on LPDR \cite{zheng2013, yoon2012, zhou2012, anagnostopoulos2006}, existing systems generally cannot recognize the plate at high accuracy while also completing LPDR in real-time. LPDR performance suffers when there is no advanced hardware to capture high-quality images or the recognition is needed for a plate attached to a fast moving car~\cite{hsu2013application}.

The difficulty lies in the extreme diversity of character patterns. Characters on the plates can be in different sizes, fonts, and colors depending on their states and nations, distorted by the viewpoint of camera, and captured with low-quality images due to lighting, shadows, occlusion, or blurring. LPDR is made even harder if there is a requirement for real-time processing. The highly complicated backgrounds introduce additional challenges, and often lead to false alarms in plate detection. Some example backgrounds include the general texts on shop boards, text-like patterns on the car windows, as well as guardrails and bricks on the road whose textures are similar to license plates.

A complete LPDR process is typically composed of two phases: license plate detection (LPD) and license plate recognition (LPR). LPD aims to identify and localize a license plate, and generate a bounding box around the plate. Usually, a verification process is also needed to guarantee that the plate is correctly detected. In LPR, plates are firstly segmented into several characters, and then each is recognized by a recognizer.

The  aim  of this work is to achieve LPDR at high accuracy in real-time, and it is critical to have LPD and LPR both efficient. To reduce the running-time in the LPD procedure, we apply the most efficient edge-based algorithm. Despite its quick processing speed, this method is sensitive to noise, and cannot deal with broken edges and remove irrelevant edges properly. The segmentation of characters is difficult if there exist faked picture borders.
Being able to learn mid and high level features from the training data,  convolutional neural networks (CNNs)  have shown good recognition performances in many vision tasks, such as image classification and object detection \cite{krizhevsky2012}. Although CNN appears to be promising in recognizing general characters, it is often difficult to obtain enough data to well train CNN for LPDR, given the diversity of plate types, the varying quality of images captured during car mobility, and the complex backgrounds around plates. This will cause a CNN to overfit easily. Finally, existing LPDR systems often take separate and independent algorithms for the candidate verification~\cite{muhammad2016,yang2015}, which introduces additional computation cost.

\begin{figure}[!t]
\setlength{\belowcaptionskip}{-10pt}
\centering
\includegraphics[width=0.48\textwidth]{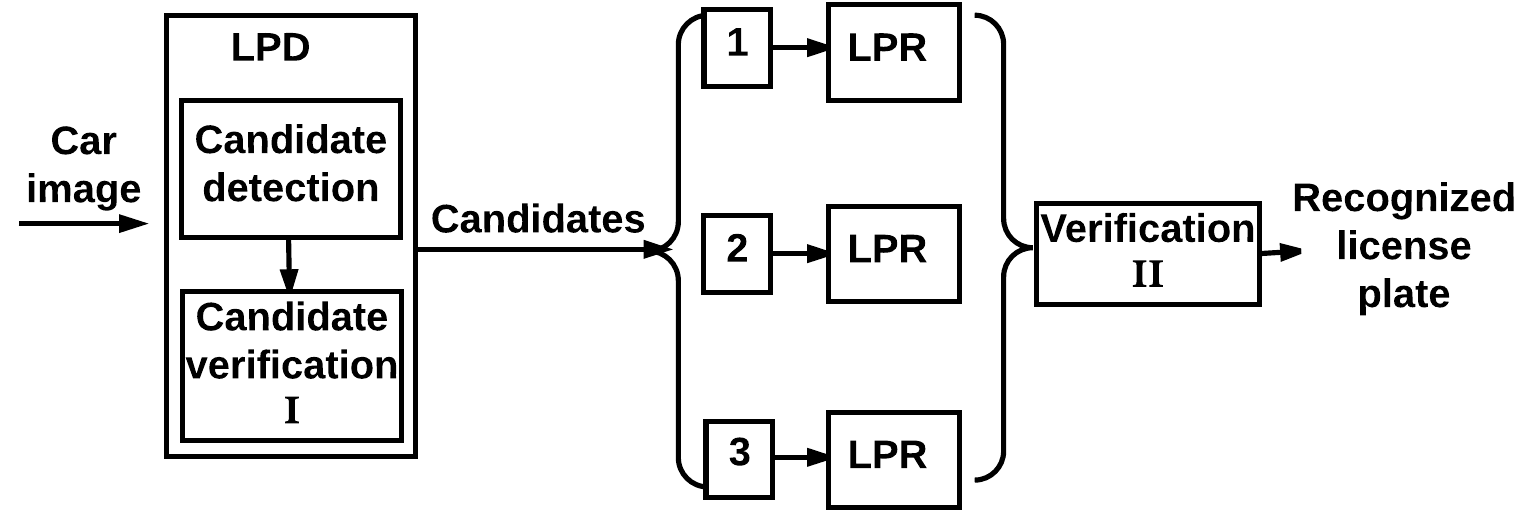}
\caption{The LPDR system structure.}
\label{pipeline}
\end{figure}

To address these issues, we implement a complete LPDR system to enable real-time and accurate license plate detection and recognition, with the structure shown in Fig.~\ref{pipeline}. The contributions of our work are as follows:

\begin{enumerate}

\item We propose a \textit{contour reconstruction} scheme to quickly localize the plate in the presence of incomplete or distorted plate contours in a practical environment.


\item We refine the license plate candidates with a simple \textit{zero-one-alternation} method to effectively remove the irrelevant edges and noise.


\item  We introduce a few strategies in our design to overcome the overfitting problem for more accurate character recognition, including data augmentation, incorporation of SIFT features to the network, and transfer learning to determine the initial training parameters for the network.


\item We propose a two-phase verification method to determine the correct plate: 1) We first apply a {\em statistical filter} in the LPD stage to effectively remove the wrong plates from the candidate set to reduce the processing overhead for further processing in a LPDR system; and 2) We take advantage of our accurate character recognizer to help verify the candidates after the LPR process. Without need of extra technique and time to verify the plate, the second phase is equivalent to shortening the system pipeline, which further improves the efficiency of our LPDR system.


\end{enumerate}

The paper is structured as follows: we introduce the related work in Section \ref{related_work}. We introduce the LPD and LPR problems, and our methods and CNN model to address the issues in Section \ref{lpd} and Section \ref{lpr}. The experimental results are presented and analyzed in Section \ref{experiment}. Finally, we conclude the paper and introduce our future work in Section \ref{conclusion}.

\section{Related Work} \label{related_work}
We briefly introduce the previous work which targets for different phases of the LPDR system.

\subsection{License Plate Detection}
For the license plate detection (LPD), previous studies \cite{anagnostopoulos2008license, du2013} generally try to capture certain morphological, color or textural features of a license plate. They are either computationally expensive and thus not suitable for real-time systems, or very easy to be affected by the color change in plates. Hough transform methods \cite{le2006} assume that license plates are defined by lines around them, and require a large memory space and considerable amount of computing time. Histogram-based approaches \cite{nejati2015} do not work properly on the images with big noise or license plates tilted. Learning methods with sliding window \cite{anagnostopoulos2006,zheng2013} suffer from high computational cost with their  applying the classifier to a sequence of rectangles within an image. In addition, not all objects are box-shaped, and representations may be polluted by features not belonging to the object. Edge-based approaches are the simplest and fastest \cite{ai2013, jiao2009}, and are employed in our work.
\cite{laroca2018robust,wang2022,usama} proposed to use Depthwise Separable Convolution Networks or YOLO Detector for the License Plate Detection task.
However, existing methods are sensitive to unwanted edges often appearing close to the license plates, which may lead to a wrong detection. We propose a simple method which exploits \textit{contour reconstruction} together with \textit{statistical filtering} to quickly and accurately detect the license plate. Different from the use of global search such as with sliding windows, our design follows the nature of human visual detection, where an attention is driven to certain locations by low-level features of images, such as contours, edges, and texts, rather than uniformly to all locations.

\subsection{License Plate Recognition}

The License plate recognition (LPR) step consists of three parts: preprocessing, segmentation, and character recognition.

In preprocessing, skew correction and image refinement are  often applied to deskew the image and remove the unnecessary borders and noises. Then a projection-based method is applied for segmentation \cite{ai2013}, which often does not work well when there are redundant borders and other edges around the license plate, especially when the faked borders appear at the top or bottom part of the plate. Based on the character features of license plates, we propose a \textit{zero-one-alternation} method to correctly remove the unwanted borders for more accurate segmentation.

Character Recognition (CR) in a general context has been widely studied. The template matching method~\cite{zheng2013} is simple and straightforward, but is vulnerable to font, rotation, noise, and thickness changes. SVMs \cite{dong2014} and shallow BP neural networks \cite{laxmi2014license} are also popular, but they are not good enough to get the most important information from the characters.

CNN-based methods have been proven to be very efficient in image recognition and classification with their ability of learning richer and higher level representations of features \cite{krizhevsky2012,yang-etal-2019-latent,yang2018cross,yang2014geo,feng2015,heming_fashion,xuewen_emnlp21,xuewen_mm20,xuewen_reformer,yingru_aaai,yingru_sde}. They are invariant under small shifts, distortions, and noise. Despite the potentials, CNNs need a huge amount of training data to avoid ``overfitting'',  while it is often very hard to get a big training set in real-world applications. With the large variety of license plates and their differences in varying environment conditions, getting enough data for training becomes even harder. 
This will significantly compromise the CR performance.  To improve the CR accuracy, we propose various schemes for data augmentation, aggregate SIFT features, and exploit transfer learning.



\section{License Plate Detection} \label{lpd}

License plate detection (LPD) is the first critical step of the LPDR system. Our LPD consists of two major procedures: candidate plate detection and candidate plate verification. Candidate plate is defined as the area that potentially contains a license plate. To more effectively detect the candidate image region of the plate, we propose a set of schemes based on edge detection, so that the scheme can better work under different backgrounds, motion blur, light conditions, and tilt angles.

\subsection{Candidate Plate Detection}
With the edge detection, some of the edges may form closed contours, and one may be around the license plate. A license plate is generally bounded by a rectangle area which contains some texts. However, in a practical environment, the contour of the license plate may not be a rectangle. It may be broken or tilted (skewed). In order to improve the detection accuracy, we first propose to reconstruct the edges to form complete rectangular candidate contours that possibly contain the plate. To speed up the detection process, we further propose a set of schemes to reduce the candidate set.


As an intuitive way of finding the location of a license plate, one can first get rectangular contours after the edge detection, and then select the one with the right width, height, and convex area. This method suffers when the contours are broken or greatly tilted. It also cannot be applied to detect the license plates with a wide range of height, width, and area.



\begin{figure}[!t]
\setlength{\belowcaptionskip}{-10pt}
\begin{center}$
\begin{array}{ccc}
\centering
\includegraphics[scale=.18]{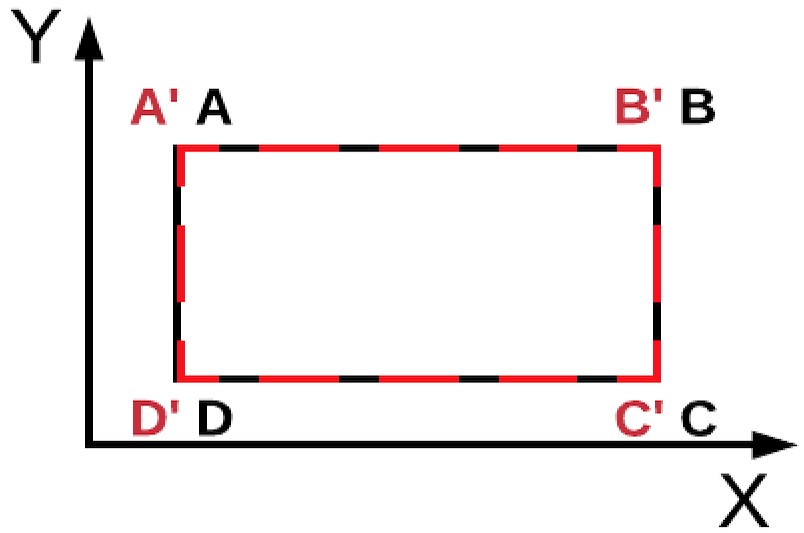} &
\includegraphics[scale=.18]{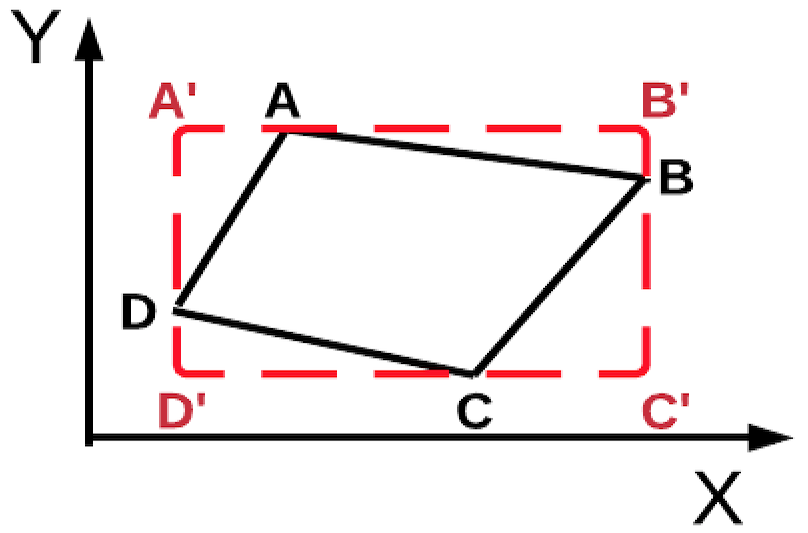} &
\includegraphics[scale=.18]{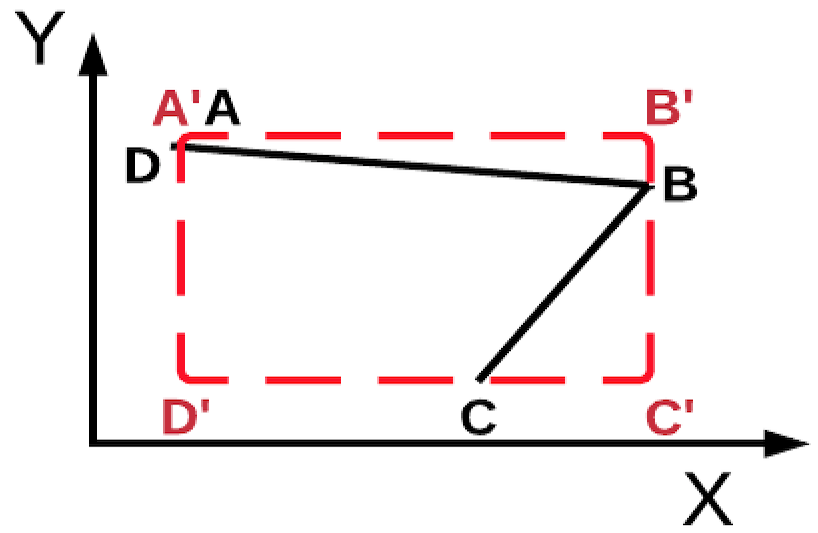} \\
\small (a) &  (b) & (c) 

\end{array}$
\end{center}
\caption{Candidates of different scenarios and the corresponding reconstruction results: (a) a normal candidate. (b) A tilted candidate. (c) A tilted broken candidate}
\label{reconstruct}
\end{figure}

To better deal with the plate candidates with broken or highly tilted contours, we propose to reconstruct a complete contour around the plate based on the extreme points found on the edge map.
As shown in Fig.~\ref{reconstruct}, if we have the top-most point $A$, the right-most point $B$, the bottom-most point $C$, and the left-most point $D$, we can reconstruct the candidate contour by taking the rectangle $A^\prime$$B^\prime$$C^\prime$$D^\prime$, where
\begin{math}
x_{A^\prime} = x_D, y_{A^\prime} = y_A, x_{B^\prime} = x_B, y_{B^\prime} = y_A, x_{C^\prime} = x_B, y_{C^\prime} = y_C, x_{D^\prime} = x_D, y_{D^\prime} = y_C.
\end{math}

With this procedure, no matter the contour is broken, tilted, or concave, we can always reconstruct a proper rectangle around it. For example in Fig.2(c), a broken contour with only two borders is still able to be reconstructed. Although simple, our performance studies indicate that this process helps to significantly improve the accuracy of LPD.


\subsection{Coarse Candidate Plate Verification}
After the detection process, there may exist multiple candidate plates. To reduce the later processing overhead, we will filter out the wrong candidates and keep only the ones most likely to be correct. We propose a {\em Statistical Filtering} method in this section.

Before presenting our scheme, we first introduce a vectorization process to translate a plate image containing alphanumeric characters to a ``pixel vector'' by summing  up the ``pixel matrix'' along its columns and then normalizing it by 255:
\begin{math}
\textbf{v} = \frac{\sum_{i}\textbf{M}_{ij} }{255}.
\end{math}

\begin{figure}[!t]
\setlength{\belowcaptionskip}{-10pt}
\centering
\includegraphics[width=0.48\textwidth]{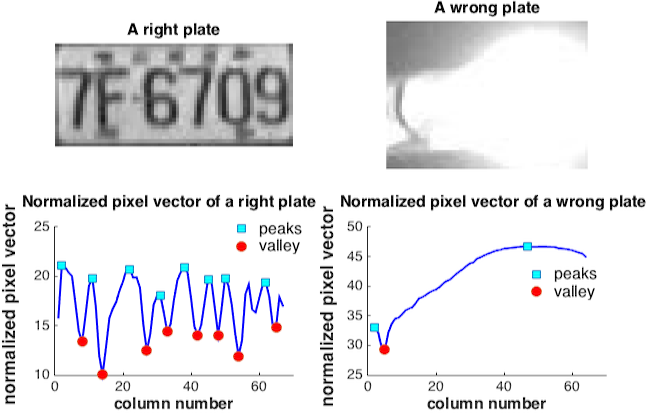}
\caption{Candidates and their normalized pixel vectors.}
\label{candidates}
\end{figure}

We observe that the pixel vector exhibits certain \textit{statistical regularities} distinguishable from the background. In the plots of the plate candidates and their 1-D pixel vectors on Fig.~\ref{candidates}, we can see that a true candidate has some distinctive statistical regularities different from the false ones. For example, a true license plate generally has a sequence of ``peaks'' and ``valleys'', and the number of peaks is larger than most of the false ones.

\begin{definition}
A value is considered to be a ``\textbf{peak}'' if it is a local maximum of the pixel vector and the other minima points on its left and right are smaller than it by a threshold.
\end{definition}

A similar definition can be applied to ``valley''. In the Fig.~\ref{candidates}, suppose the threshold is $3$, then there are $8$ peaks on the correct plate, while  there are only $2$ peaks on the wrong plate. A candidate with more than 6 peaks (the number of characters in the plate) will be considered as a potential plate and passed to the subsequent procedures. The others will be eliminated.

\section{License Plate Recognition} \label{lpr}

License plate recognition (LPR) consists of three steps: pre-processing, segmentation, and character recognition (CR). Pre-processing procedures such as skew correction and image binarization are commonly taken first to prepare an image for the subsequent steps. A refining procedure is then applied to remove the unnecessary parts (e.g., the bordering regions) from the binary image, so it will ideally look like that in Fig.4(b). Finally, with the binary image in Fig.4(b), a pixel vector can be built, as shown by the plot in Fig.4(e). The segmentation is performed along the red dotted lines passing through the valleys of the 1-D vector. This segmentation method is called \textit{vertical projection}. The segmented letters are shown in Fig.4(f). With the existence of noisy borders around the plate, the segmentation step is often difficult. Although the left and right borders can be removed through the literature method \cite{yoon2012}, existing methods often fail to remove the top and bottom borders in the presence of distracting image contents. This may lead the segmentation error. We propose a simple  {\em zero-one-alternation} (ZOA) method which can effectively and quickly remove the top and bottom borders. After the characters are segmented from the license plate, a robust CR method that can work in different situations is followed to better recognize the segmented characters.
\begin{figure}[!t]
\setlength{\belowcaptionskip}{-10pt}
\begin{center}$
\begin{array}{cc}
\centering
\includegraphics[scale=.23]{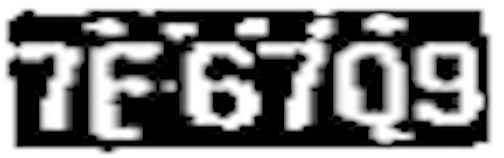} \label{seg:a} &
\includegraphics[scale=.23]{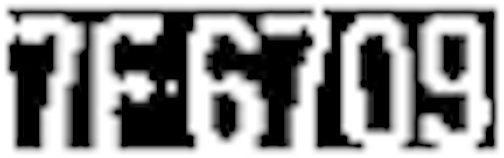} \label{seg:b} \\
(a) & (b) \\
\includegraphics[scale=.23]{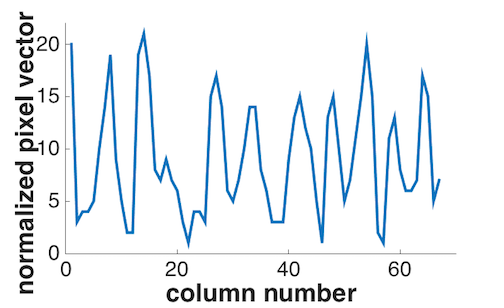} \label{seg:c} &
\includegraphics[scale=.23]{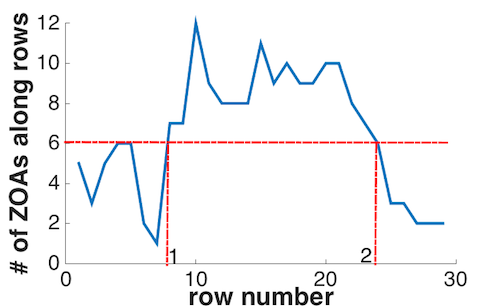} \label{seg:d} \\
(c) & (d) \\
\includegraphics[scale=.23]{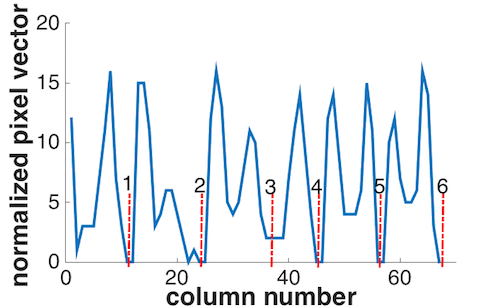} \label{seg:e} &
\includegraphics[scale=.23]{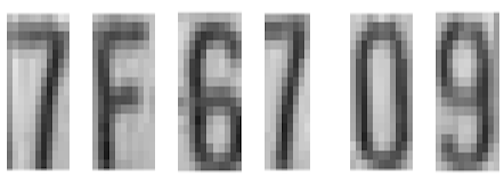} \label{seg:f} \\
(e) & (f) \\
\end{array}$
\end{center}
\caption{Plate preprocessing and segmentation illustration: (a) The binary plate. (b) The refined binary image after ZOA processing. (c) Normalized pixel vector for (a). (d) Number of zero-one-alternations along rows. (e) Normalized pixel vector for (b). (f) The segmented characters.}
\label{fig:seg}
\end{figure}


\subsection{Refinement with Zero-One-Alternation}

A binary image after the pre-processing has two values, 1 (255) or 0. To more effectively remove the noisy top and bottom borders outside the characters of the license plate, we propose a novel scheme called {\em zero-one-alternation} (ZOA).

\begin{definition}
Along each row of the binary image matrix, the number of changes from 0 to 1 or from 1 to 0 is the number of \textbf{zero-one-alternations} (ZOAs).
\end{definition}

By scanning row-by-row from the top to the bottom of the binary image, if we can observe similar number of ZOAs (more than $\alpha$) from row $i$ until row $i+n$, with $n \geq \beta$, it is very likely that the characters are contained within a region of $n$ rows. In Fig.4(d), this region is from point $1$ to point $2$. $\alpha$ and $\beta$ are the only parameters need to be predefined. $\alpha$ and $\beta$ can be set by the smallest number of alphanumeric characters and the lowest height of the characters in pixels in all the license plates. In this work, we set $\alpha=6$ and $\beta=8$. Note that these two are heuristic parameters, but should work well in most cases.


After removing the top and bottom borders using ZOA, we get Fig.4(b). Without this step, the plate might be mistakenly recognized as ``7E67Q9'' instead of ``7F6709''. To remove the left and right borders, we further apply a commonly used method in \cite{yoon2012}. The segmentation can be efficiently performed over the properly trimmed image through the vertical projection  introduced earlier.

\subsection{Character Recognition}

The last key phase of a LPDR system is to recognize the segmented characters of the license plate. It is necessary and important to propose a robust CR method that can work in different situations.

Due to the large variety of license plates and their images differ in varying environment conditions, there is no specific dataset for plate character recognition. Thus, we create a dataset by ourselves, as introduced in Section \ref{dataset}. However, the training data available are far from enough. This will significantly compromise the CR performance. In our created dataset, we have 36 classes and 3,000 examples in total. So on average less than 100 training data are available for each class, where a typical classification problem requires more than 1,000 training data for each class. Thus, the character recognition for license plates is challenging. With very limited training data, the learning faces the problems of \textit{over-fitting}, where the trained model does not apply well to new data.

To enable higher quality CR for LPDR system, we will exploit the use of data augmentation, a newly designed SIFT-CNN model, and the transfer learning.

\subsubsection{Data Augmentation}

In order to make full use of the limited training samples, we consider two methods for data augmentation. In the first method,  we augment data via a number of random affine transforms to avoid the duplication of data in the training set.
We gradually increase the amount of augmentation till we have a low testing error. This helps to alleviate the overfitting problem and make our model more easily applied to new scenarios.

A big challenge of performing accurate CR for license plate is that a plate image is often subject to different illumination and noise. In the second method, we vary the illumination of our training data and inject Gaussian noise into our model so that it performs more robustly under different image quality. Our augmentation parameters are summarized in table \ref{table_aug}.

\begin{table}[!t]
\centering
\begin{tabular}{|c|l|}
\hline
\bfseries Technique & \bfseries Parameters\\
\hline
Random rotation & $\pm 10$ degrees\\
\hline
Random translation & $\pm 5$ pixels in any direction\\
\hline
Random zoom & factors of 1 and 1.3\\
\hline
Random shearing & $\pm 25$ degrees\\
\hline
Inverting intensities & 255 - \textit{original intensities}\\
\hline
Gaussian noise & (mean, deviation) = (0, 0.1)\\
\hline
Illumination change & $\pm 20\%$ image intensity\\
\hline
\end{tabular}
\caption{Data augmentation setup}
\label{table_aug}
\vspace{-0.3in}
\end{table}

\subsubsection{Incorporating SIFT feature vector into CNN}

CNN requires massive training data to work well in different scenarios. As an another scheme, scale-invariant feature transform (SIFT) method \cite{Lowe1999} can extract image descriptors for recognition without requiring a large amount of data. SIFT searches an image at multiple scales and positions to look for the regions that have the maxima or minima with high contrast, called keypoints.
The image gradients can be tracked with a histogram and represented as a descriptor to characterize the appearance of a keypoint. To describe a keypoint, the region around the keypoint is divided into $4 \times 4$ subregions, within each gaussian derivatives are computed in 8 orientation planes. So a 128-dimension descriptor vector is formed for each keypoint, and this procedure is repeated for all keypoints to obtain a set of vectors for the image. SIFT descriptors are invariant to translations, rotations, and scaling of the image, and robust to moderate transformations and illumination variations. Despite these benefits,  SIFT filters are fixed and cannot vary for a different source of data, which compromises its performance. To take advantage of both schemes while avoiding their limitations, we propose a new SIFT-CNN model which integrates the vector extracted by SIFT with the feature vector of CNN to increase the accuracy of character recognition.



Our method is inspired by a \textit{bag-of-words} (BoW) representation where image features are treated as words. In BoW, detection and description of image features are first applied, followed by assigning feature descriptors to a set of predetermined clusters (like vocabulary). Finally, a vector called a BoW is used to track the number of occurrences of words in the vocabulary.



\begin{figure}[t]
\setlength{\belowcaptionskip}{-10pt}
\centering
\includegraphics[width=0.48\textwidth]{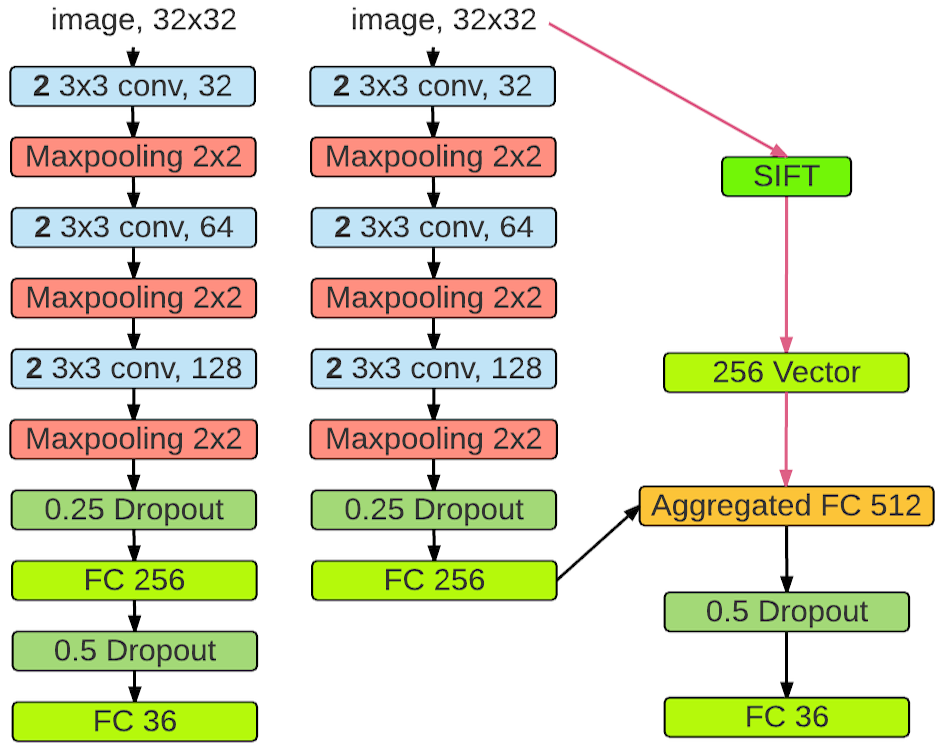}
\caption{Network architectures for CR problem. \textbf{Left:} traditional CNN-8 model as a reference. \textbf{Right:} CNN-8 incorporated with SIFT (SIFT-CNN) model.}
\label{sift_cnn}
\end{figure}

Fig.~\ref{sift_cnn} shows our SIFT-CNN model, where a SIFT branch is added to extract the image feature vector as follows:
\begin{enumerate}
 \item Using SIFT method to extract the keypoint descriptors of the image.
\item Assigning each keypoint descriptor to a set of clusters whose centers are determined using K-means based on the descriptors of training data. 
\item Constructing the feature vector $\mathbf{v}$ for each image to track the number of keypoints assigned to each cluster, and normalizing $\mathbf{v}$ by $\frac{\mathbf{v}}{\vert \mathbf{v} \vert_{l_2}}$.
\item Concatenating the feature vector $\mathbf{v}$ with the feature vector extracted in the fully-connected layer of CNN to form a final feature vector.
\end{enumerate}



We set the number of clusters to 256, so that the length of the feature vector is the same as that of the CNN feature vector. Then the concatenated feature vector has the size 512, and will be used to classify the image.

\subsubsection{Efficient Learning across Applications}

Many machine learning methods work well only under a common assumption, the training and testing data are taken from the same feature space and follow the same distribution \cite{pan2010}. When the distribution changes, most statistical models need to be rebuilt using newly collected training data. It would be helpful to exploit an emerging technique called the transfer learning, which extracts the knowledge from one or more source tasks and applies the knowledge to a target task. For a small training dataset, without enough knowledge on the data, randomly initializing a neural network can make the training result worse since it is easily to get stuck in local minima.  However, if a neural network can start from an already trained feature extractor, it can \textit{``borrow''} some knowledge from a task already learnt to the new task. Despite the potential, it is crucial and difficult to determine the right knowledge sources and the amount of knowledge to use for a new learning task. If too few are transferred, we may not get enough knowledge from source tasks, while transferring too many parameters will not only lead to redundancy but may also compromise the training performance if many parameters are irrelevant.

\begin{figure}[!t]
\setlength{\belowcaptionskip}{-10pt}
\centering
\includegraphics[width=0.48\textwidth]{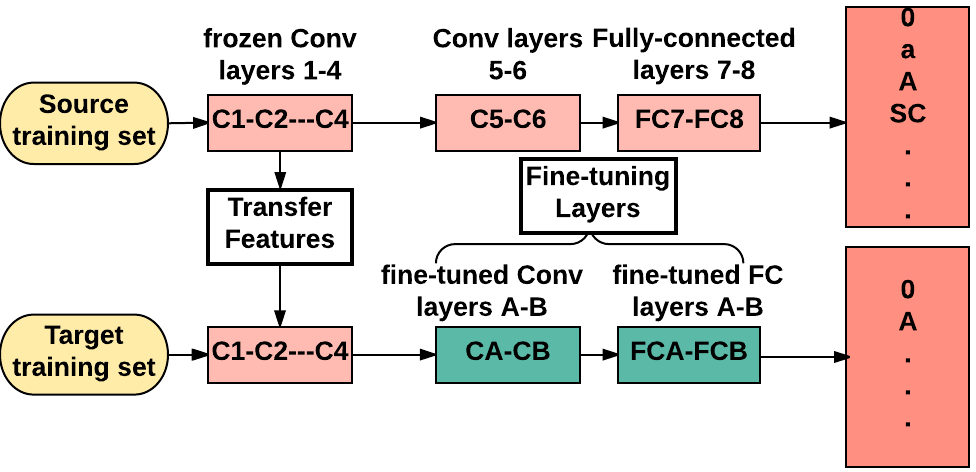}
\caption{Hybrid CNN model. Note: SC-Special Characters.}
\label{transfer2}
\end{figure}

To further improve the quality of character recognition, we apply  transferring learning to the training of our CNN model. Our preliminary studies indicate that features learnt from the low-level and mid-level of a neural network are more effective for transfer learning.
We \textit{``borrow''} the parameters of a SIFT-CNN model, trained on the dataset ICDAR \cite{karatzas2015}. The dataset contains more than 62 classes, which include 26 upper-case letters, 26 lower-case letters, 10 digits classes and other classes. Our studies indicate that this dataset is  excellent in extracting low and mid level features of the document letters. Initially, we attempted to apply the model trained with this data set directly to recognizing the characters on the license plates. However, the CR accuracy is only  84\%, since the license plate characters are quite different from the document letters in fonts, types and classes. Instead, as shown in Fig.~\ref{transfer2}, we train a new model based on license plate data. We \textit{``borrow''} the convolutional layers $1-4$ of the trained SIFT-CNN model, and rebuild two convolutional layers $CA-CB$ and two fully-connected layers $FCA-FCB$ with the number of neurons being 512 and 36 respectively.





\subsubsection{Hybrid CNN model}

For recognizing characters on license plates, we apply  \textit{Hybrid CNN}  which exploits a SIFT-CNN model with transfer learning. We train our model using stochastic gradient descent (SGD) algorithm with an annealed learning rate. We test our model with different number of transferred layers and fine-tuned layers, and we find four transferred layers and four fine-tuned layers can achieve the best result. Thus, we fix the 4 transferred layers and only fine-tune the 4 rebuilt layers using the target training data created for this work. The training process stops when there is no improvement in performance for 5 epochs. In the experiment section, we will show that the transfer learning together with fine-tuning can improve the recognition performance, especially when the training data are very limited.

If we fine-tune all layers, it will not only take a very long time, but may also break the good feature filters already built by the original model.
We only fine-tune the last two convolutional layers and the two fully-connected layers. The features extracted from ICDAR dataset by the low-level and mid-level convolutional layers are very general, and can be easily applied to the classification of our images. The higher convolutional layers trained with the ICDAR dataset, however, do not fit well in our system and cannot provide satisfying CR results.

\subsection{Fine Candidate Verification with Voting}
After the statistical filtering in the LPD phase, we might still have more than one candidate plate, usually two or three. To determine which one is correct, all the remaining candidates should go through a second verification phase. We propose a``voting'' method taking advantage of the results from the CR process for more accurate verification without need of additional computing time and resources. As every character is recognized independently, a candidate is more likely to be a correct one if  every character is more likely to be correctly recognized. So every character is ``voting'' for the candidate they belong to.

The probability of the character $i$ to be the right label $y_i$ is
\begin{math}
P_i(y=y_i|\textbf{X})
\end{math}
, as shown in table \ref{real false}. The probability of the candidate to be the right plate is
\begin{math}
\hat{P}=\prod_{i=1}^{n} P_{i}
\end{math}
, where $n$ is the number of segmented characters. The one with a higher probability will be selected as the license plate.
From table \ref{real false}, we can tell that the first candidate is much more likely to be the right candidate than the second one since the first one has a higher probability $\hat{P}$.
Once the CR process is completed, the verification is followed to choose the right candidate and the final result is achieved.

\begin{table}
\centering
\begin{tabular}{|c|c|c|c|l|}
    \hline
    Characters & Real  &  $P_r (\%)$ & False & $P_f (\%)$\\
    \hline
    7 & $C_{r1}$   &   93.5 & $C_{f1}$ & 69.6\\
    \hline
    F & $C_{r2}$   &   94.1 & $C_{f2}$ & 50.1\\
    \hline
    6 & $C_{r3}$   &   88.6 & $C_{f3}$ & 86.8\\
    \hline
    7 & $C_{r4}$  &   93.9 & $C_{f4}$ & 42.4\\
    \hline
    0 & $C_{r5}$   &   85.9 & $C_{f5}$ & 49.4\\
    \hline
    9 & $C_{r6}$   &   99.9 & $C_{f6}$ & 56.8\\
    \hline
     plate & $\hat{P_r}$    &   62.8 & $\hat{P_f}$ & 3.6\\
    \hline
\end{tabular}
\caption{Illustration of ``voting'' on a real \& a false candidate.}
\label{real false}
\vspace{-0.2in}
\end{table}

\section{Experiments} \label{experiment}

We compare the performance of our system with several state-of-the-art LPDR schemes, and we emphasize the key similarities and differences. To demonstrate the accuracy and efficiency of our proposed LPDR scheme, we conduct a set of experiments over natural car images taken in different environments and nations. Our experiments are run on a dual-core 2.7 GHz Intel i5 machine, and our algorithms are realized in C++ due to its efficiency.

\subsection{Dataset} \label{dataset}

We test our system on two datasets. The first one is the AOLP benchmark dataset \cite{hsu2013application} with three subsets: access control (AC), law enforcement (LE), and road patrol (RP). AC refers to the cases that a vehicle passes a fixed passage with a low speed or full stop. The 681 images were captured under different illuminations and weather conditions, with the resolution of each image at $240 \times 320$. LE refers to the cases that a vehicle violates traffic laws and is captured by roadside cameras. The 757 image samples were collected with the image resolution at $240 \times 320$ or $480 \times 640$. RP refers to the cases that a camera is installed on a patrolling vehicle, and the 611 images were taken with arbitrary viewpoints and distances, with the resolution of each image at $240 \times 320$.  The second dataset, LongIsland,  contains 300 samples of car images we collect ourselves by driving on local roads of Long Island. We drove with the average speed of 30 mph on a cloudy day. The resolution of each image is $480 \times 640$. The dataset consists of car images of different backgrounds, motion blurring, skew angles, capturing distances, and sizes. The images in the dataset are labelled manually.

Two alphanumeric dataset are used in the CR phase. The ICDAR dataset \cite{karatzas2015} consists of about 12,000 samples. It includes classes for 10 digit numbers, 26 classes for upper-case characters, 26 classes for lower-case characters, and many other special symbols. We use the 6548 samples of 10 digits and 26 upper-case characters to train our CNN models. We create the second dataset ourselves by cropping the car and motorcycle license plates of different fonts, nations, and states from the Internet. It consists of 3,000 training samples of 36 different classes, which include 26 upper-case letters and 10 digits. The two datasets are merged together and then divided into three subsets - training subset ($\frac{5}{7}$ of the samples in the merged dataset), validation subset ($\frac{1}{7}$ of the merged dataset), and test subset ($\frac{1}{7}$ of the merged dataset). The training subset is used for model training, the validation subset for validating our models and enforcing early-stopping of training when the validation accuracy does not have perceivable improvement over time, and the test subset for testing the performance of our models. When training our Hybrid CNN model, we divide the two datasets in the same way. However, we first train our model on ICDAR dataset to get the initial parameters, then fine-tune the parameters with the plate character dataset we created.

\subsection{Evaluation Metrics and LDPR Results}

To measure the processing speed of our system, we calculate the average number of images processed by our system in one second, or {\em frames per second} (FPS).

The performance of plate detection is evaluated using the {\em precision} and {\em recall} rate \cite{karatzas2015}, two most widely used evaluation criteria for detecting the general texts in natural images. Precision is defined as the ratio of the correctly detected license plates and the total number of detected plates. Recall is the ratio of the correctly detected license plates and the total number of groundtruth. A license plate is correctly detected if it is totally enclosed by a bounding box, and the Intersection over Union (IoU), which is area of overlap divided by area of union, is greater than 0.5.

We evaluate the plate recognition performance with {\em recognition rate}, which is defined as the number of correctly recognized license plates divided by the total number of correctly detected plates. Note that a correctly recognized license plate means all the characters on the plate are recognized correctly.

\begin{figure}[!t]
\setlength{\belowcaptionskip}{-10pt}
\begin{center}$
\begin{array}{cc}
\centering
\includegraphics[scale=.18]{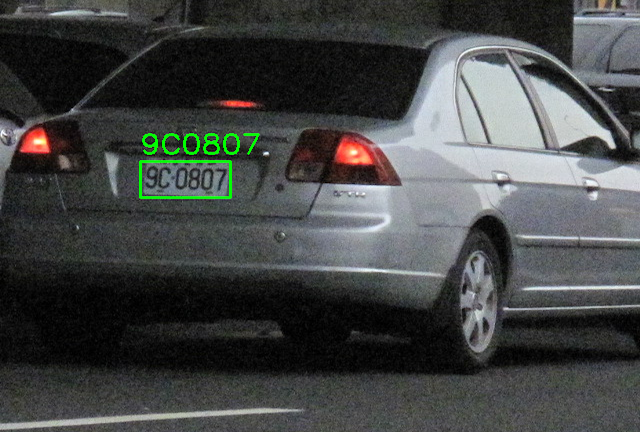} &
\includegraphics[scale=.18]{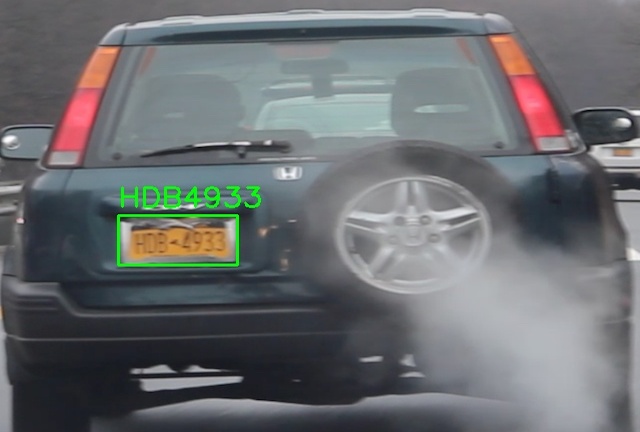} \\
\includegraphics[scale=.36]{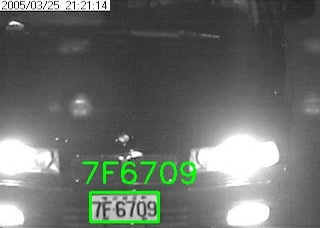} &
\includegraphics[scale=.36]{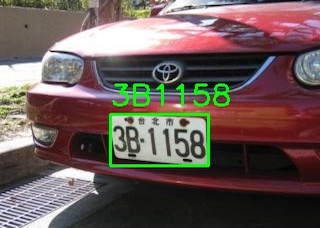} \\
\end{array}$
\end{center}
\caption{Some of the image samples from the AOLP and LongIsland datasets with the recognition results .}
\label{res}
\end{figure}

Some of the results of our system are shown in Fig.~\ref{res}. We will evaluate the detailed performance on LPD and LPR in Section ~\ref{LPD} and ~\ref{LPR} respectively, and show the impact of motion on the LPDR performance in Section~\ref{Motion}.

\subsection{Performance of Plate Detection}
\label{LPD}


Our system exploits Contour Reconstruction and Hybrid CNN for plate detection and recognition, and we call it CR-HybridCNN. Similar abbreviations are given to the other systems likewise. We compare the performance of plate detection of our system with four other methods proposed within the recent three years and showed to have good LPD and LPR performance in their papers.

The detection performance of all methods are evaluated with the AOLP dataset with the results shown in Table~\ref{table_detect_aolp}. Based on the metrics of evaluation, our proposed system is the most efficient one, and it can process 90 frames per second for images in all three subsets. While it is slightly less accurate than EC-LDA, the processing speed is about 20 times larger. Next we introduce each reference LPDR method and show the difference of our design.

\textbf{Edge-clustering and Linear Discriminant Analysis} (EC-LDA) \cite{hsu2013application} uses an Expectation-Maximization (EM) edge clustering algorithm to extract regions with dense sets of edges which have shapes similar to plates. The clusters with their edge densities larger than a predefined threshold and edges of plate-like shapes will be considered as plate candidates. As the system needs to predefine eight threshold values and parameters such as the number of clusters and the penalty weight, it prohibits the system from working well in other scenarios. In addition, the clustering algorithm is computationally expensive with its needs for a large number of iterations to obtain a good result.

Our system replaces all of these computationally expensive part with only one simple reconstruction method, which only needs two general thresholds without other parameters. Thus our algorithm has its accuracy comparable to EC-LDA, but can work much faster system and better in new scenarios than EC-LDA.

\textbf{Extremal Regions and RBM} (ER-RBM) \cite{gou2016vehicle} applies morphological transformations with several rounds of close and open operations to do coarse detection and produce the candidate license plates. Then Extremal Regions method is used to detect the license plates at the fine level. Each stage of this complex pipeline must be precisely tuned independently, so the system is slow, and it takes more than 0.2 seconds on average to process an image.

Similar to ER-RBM, our system also tries to first coarsely detect license plates and then detect the final results by comparing several candidates. However, our system uses the simple { \em statistical filtering} method to complete the coarse detection quickly, and reduces candidates to a small number (usually fewer than three) in one round rather than several rounds of transformations. In addition, in the fine detection stage, we exploit the results from character recognition to vote for the final result without additional processing. Thus our processing speed is more than 10 times faster.

\textbf{Other Real-Time LPDR Systems}
Many research efforts in LPDR focus on speeding up the detection pipeline. However, after testing on our own machine, only KNN-SVM~\cite{tabrizi2016hybrid} and Color with Edge Detection and KNN (CE-KNN) \cite{qiu2016optimized} can run in real-time (with 30 FPS or better). However, their detection accuracy is sacrificed. KNN-SVM combines edge detection with morphological transformation for plate detection. CE-KNN uses color detection and edge detection to process the image separately, and then uses the information from the two to locate the final license plate. Both of the two methods need to limit the size of the candidates to be within a small range, so that the final candidate within the range is considered as the plate. If the sizes of license plates vary over a large range, the system's performance will suffer. Our license plate detector is general and does not depend on the application scenarios, dataset, and sizes of license plates.


\begin{table*}[!t]
\centering
\begin{tabular}{|l|*{10}{c|}}
\hline
\multirow{2}{*}{Methods} & \multicolumn{3}{c|}{Subset AC} & \multicolumn{3}{c|}{Subset LE} & \multicolumn{3}{c|}{Subset RP} \\\cline{2-10}
& Precision & Recall & FPS & Precision & Recall & FPS   & Precision  & Recall & FPS
\\\hline
CE-KNN    & 0.785 & 0.794 & 90 & 0.778 & 0.787 & 85 & 0.755 & 0.779 & 85
\\\hline
KNN-SVM & 0.796 & 0.792 & 89 & 0.784 & 0.792 & 85 & 0.773 & 0.785 & 79
\\\hline
ER-RBM   & 0.842 & 0.853 & 9 & 0.848 & 0.851 & 7 & 0.822 & 0.828 & 6
\\\hline
EC-LDA  & 0.925 & \textbf{0.948} & 6 & \textbf{0.924} & \textbf{0.939} & 5 & \textbf{0.918} & \textbf{0.927} & 3
\\\hline
Proposed & \textbf{0.939} & 0.913  & \textbf{100} & 0.922 & 0.909  & \textbf{95} & 0.909 & 0.894  & \textbf{90}
\\\hline
\end{tabular}
\caption{Comparing the detection accuracy and speed of  plate detectors. Precision, recall, and frames per second (FPS) are used as evaluation metrics. Our system has the fastest detector while still ensuring its detection precision and recall rate comparable with the most accurate system EC-LDA, with its detection precision and recall less than 2\% lower on average.}
\label{table_detect_aolp}
\end{table*}

\subsection{Performance of Plate Recognition}
\label{LPR}

In EC-LDA, local binary pattern (LBP) features are extracted and classified using a two-layer LDA classifier. The training samples are randomly selected from the three subsets and processed by the EC-LDA system. This random selection of training data cannot prevent the use of same data for testing, which may compromise the effectiveness in performance evaluation.

In ER-RBM, a restricted Boltzmann machine is used as the classifier. The results from experiments of the paper show that RBM performs better than SVM in classifying plate characters.

In KNN-SVM, KNN is used as the initial step to classify all datasets and then multi-class SVM is performed only over the smaller dataset with similar characters. It performs better than traditional SVM classifiers.

CE-KNN proposes to use normalization, image thinning, and feature extraction as pre-processing. The extracted features include the slope of stroke, the amplitude of inflection point, and the depth of profile. Then the features are used to train the classifier.

In our system, we use deep convolutional neural networks, which have been tested to work well in many computer vision tasks \cite{krizhevsky2012}. We also apply several data augmentation techniques to increase training data, and exploit the transfer learning in our SIFT-CNN model to first classify general upper-case letters and digits. The designed network is fine-tuned with training over additional plate characters in the dataset created by our own.

\subsubsection{Effectiveness of the Hybrid-CNN Model}

The character recognition model plays an important role in plate recognition. The Hybrid-CNN model used in our CR part is built on an eight-layer CNN model, with the incorporation of data augmentation and SIFT feature vectors, and the application of transfer learning to set up the initial parameters for training. We denote the original model without extra techniques as ``CNN'', the model using data augmentation as ``Aug-CNN'', and the model with data augmentation and SIFT vectors as ``SIFT-CNN'', based on which, ``Hybrid-CNN'' is created with the addition of transfer learning. As shown in Table~\ref{fig:cnn_classification}, with use of these three techniques, the classification accuracy improves from 0.848 to 0.892, 0.918, and 0.964, respectively.

\begin{table}[!t]
\small
\centering
\begin{tabular}{|c|c|c|c|l|}
\hline
Models & CNN & Aug-CNN & SIFT-CNN & Hybrid-CNN\\
\hline
Accuracy & 0.848 & 0.892 & 0.918 & \textbf{0.964} \\
\hline
\end{tabular}
\caption{Classification performance of different CNN models on 36 characters.}
\label{fig:cnn_classification}
\end{table}




\subsubsection{Comparison with other Schemes on LPR}

The recognition results of all methods on AOLP dataset are presented in Table \ref{table_recog}. Compared to other schemes studied, our system achieves the highest FPS for all of the three subsets. It has the highest recognition rate on the Subset AC. For other two subsets, its recognition rates are only slightly lower, with less than 1\% below the EC-LDA system. The results demonstrate that our method can perform well on different datasets.

\begin{table}[!t]
\small
\centering
\begin{tabular}{|l|*{7}{c|}}
\hline
\multirow{2}{*}{Method} & \multicolumn{2}{c|}{Subset AC} & \multicolumn{2}{c|}{Subset LE} & \multicolumn{2}{c|}{Subset RP} \\\cline{2-7}
& RR  & FPS & RR & FPS   & RR & FPS
\\\hline
CE-KNN     & 0.833 & 84 & 0.812 & 77 & 0.788 & 81
\\\hline
KNN-SVM & 0.856 & 79 & 0.824 & 70 & 0.812 & 72
\\\hline
ER-RBM       & 0.872 & 50 & 0.854 & 44 & 0.835 & 46
\\\hline
EC-LDA & 0.931 & 35 & \textbf{0.892} & 30 & \textbf{0.902} & 30
\\\hline
Proposed & \textbf{0.935} & \textbf{95} & 0.884 & \textbf{90} & 0.895 & \textbf{92}
\\\hline
\end{tabular}
\caption{Recognition rates (RR) and frames per second (FPS) of different license plate recognizers. Our system is the fastest one while also achieving the recognition rate comparable with the most accurate system EC-LDA.  }
\label{table_recog}
\end{table}

\subsection{License Plate Detection and Recognition with Motion}
\label{Motion}

It's easier to have a good performance on a fixed dataset with all the parameters well tuned. However,  both the detection and recognition become harder when cars are on move. We further compare our system with others using the LongIsland dataset, with images captured during driving.

Table \ref{table_detect_li} shows that our system achieves the highest accuracy and FPS for both plate detection and recognition, which demonstrates its robustness. EC-LDA has high detection precision, recall, and recognition rate on the AOLP dataset, but its performance drops off considerably when applied to the LongIsland dataset. Part of the drop comes from the difficulty of detecting and recognizing license plates with motion blur. The difficulty of fine-tuning the large number of parameters also affects its performance. A small parameter shift in one step may compromise the ones finely tuned in the previous steps. Its use of training data from the same dataset is another factor that prevents it from working well in other scenarios.

\begin{table}[!t]
\centering
\begin{tabular}{|l|*{4}{c|}}
\hline
\multirow{2}{*}{Method} & \multicolumn{2}{c|}{Detection} & \multirow{2}{*}{RR} & \multirow{2}{*}{FPS}  \\\cline{2-3}
& Precision & Recall & &
\\\hline
CE-KNN      & 0.623 & 0.643 & 0.815 & 38
\\\hline
KNN-SVM  & 0.655 & 0.67 & 0.847 & 33
\\\hline
ER-RBM     & 0.724 & 0.747 & 0.882 & 3
\\\hline
EC-LDA  & 0.813 & 0.82 & 0.871 & 2
\\\hline
Proposed & \textbf{0.843} & \textbf{0.827}  & \textbf{0.9} & \textbf{47}
\\\hline
\end{tabular}
\caption{Quantitative results on LongIsland dataset. Our system has the fastest detection and recognition speed, and is also more accurate than other systems. Note: RR-Recognition Rate.}
\label{table_detect_li}
\end{table}


\section{Conclusion}  \label{conclusion}

We have introduced a license plate detection and recognition (LPDR) system and demonstrated its accuracy and efficiency under different conditions. The accuracy and efficiency are achieved with a set of schemes we propose: an efficient license plate detection algorithm with the support of contour reconstruction, a refinement method based on zero-one-alternation to effectively remove the unwanted boarders for more accurate character segmentation, an efficient hybrid-CNN model along with various techniques to overcome the overfitting problem, and two-phase verification to determine the correct plate at low cost.
In our future work, we plan to apply our algorithms to LPDR in videos and make the pipeline shorter, which can be done by removing segmentation part in the plate recognition step to further improve the performance of the system.


\bibliography{example_paper}
\bibliographystyle{icml2019}

\end{document}